\documentclass[11pt, a4paper, onecolumn, copyright, goog]{google}

\usepackage[authoryear, sort&compress, round]{natbib}

\usepackage[utf8]{inputenc} 
\usepackage[T1]{fontenc}    
\usepackage{hyperref}       
\usepackage{url}            
\usepackage{booktabs}       
\usepackage{amsfonts}       
\usepackage{nicefrac}       
\usepackage{microtype}      
\usepackage{xcolor}         
\usepackage{pifont}
\usepackage{colortbl}                     

\definecolor{mazefill}{HTML}{FBE2E5}      
\definecolor{nonofill}{HTML}{DCF3DE}      
\definecolor{oursfill}{HTML}{EDEDED}      

\usepackage{amsmath}
\usepackage{amssymb}
\usepackage{mathtools}
\usepackage{amsthm}
\usepackage{graphicx}
\usepackage{subcaption}
\usepackage{hyperref}
\usepackage{algorithm}
\usepackage{algorithmic}
\usepackage{xspace}
\usepackage{multicol}
\usepackage{multirow}
\usepackage{array}
\definecolor{airforceblue}{rgb}{0.36, 0.54, 0.66}
\definecolor{turkishrose}{rgb}{0.71, 0.45, 0.51}
\definecolor{lavenderpurple}{rgb}{0.59, 0.48, 0.71}
\hypersetup{colorlinks=true, citecolor=airforceblue, linkcolor=turkishrose, urlcolor=lavenderpurple}

\newcommand{\x}{\mathbf{x}}

\newcommand{\z}{\mathbf{z}}

\newcommand{\cat}{\mathrm{Cat}}

\newcommand{\at}{\alpha_{t}}

\newcommand{\mask}{\boldsymbol{m}}
\newcommand{\method}{\texttt{CO\textsubscript{2}Jump}\xspace}

\def\lossnelbo{{\mathcal{L}^{\infty}_{\text{NELBO}}}}

\newcommand{\Eqn}[1]{Eq.~(\ref{#1})}
\newcommand{\Rt}{\mathbf{R}_t}

\uselogo{} 

\title{\LARGE{Concurrent Image Understanding and Generation: Self-Correcting Coupled Markov Jump Processes}}

\correspondingauthor{qqdd@google.com}

\reportnumber{0001} 

\author[1,3,$\dagger$]{Minh-Quan Le}
\author[1]{Armand Comas}
\author[1]{Alexandros Lattas}
\author[1]{Stylianos Moschoglou}
\author[2]{Pedro Vélez}
\author[2]{Amit Raj}
\author[1]{Aaron Germuth}
\author[1]{Thabo Beeler}
\author[3]{Dimitris Samaras}
\author[1]{Di Qiu}

\affil[1]{\thepa{}{}}
\affil[2]{Google DeepMind}
\affil[3]{Stony Brook University}

\begin{document}
\begin{abstract}
\vspace{-5mm}
Human cognition does not separate understanding and generation. A teacher at a whiteboard speaks and draws \emph{together}, each modality reshapes the other. In this paper, we bring this coupled loop to artificial systems. Masked Diffusion Models (MDMs) are ideally suited to this task, yet existing samplers either decode text and image interleavedly or independently update them in parallel branches that share only previous-step history, but not the other modality's latest decisions \emph{within} the same step; combined with MDMs' inability to remask, cross-modal contradictions are neither detected nor repaired. We introduce \textbf{Self-Correcting Coupled Markov Jump Processes (SC-CMJP)}, a framework in which one modality's transition rates are functionals of the other modality's confidence score, as weighted by cross-modal attention. Furthermore, a remasking jump retracts commitments the moment cross-modal evidence turns against them. In conjunction with SC-CMJP, we introduce \method{} (Self-\underline{CO}rrecting \underline{CO}upled \underline{Jump}), a novel training-free single-pass sampler for joint multimodal geneneration. For training and evaluation purposes, we have created and will release three large-scale joint multimodal generation corpora: \textsc{JEdit-1M}, \textsc{JMaze-200K}, \textsc{JNono-200K}, with matching in- and out-of-distribution benchmarks. \method{} achieves best joint performance for image understanding and editing as well as visual reasoning (maze and nonogram solving). The performance of the sampler scales monotonically with the number of denoising steps, evidence that the benefits of cross-modal coupling \emph{compound} across the trajectory. 
Project page: \href{https://coupled-jump.github.io}{https://coupled-jump.github.io}
\vspace{-3mm}
\end{abstract}
\maketitle
\begin{figure*}[h]
  \centering
  \includegraphics[trim=0 150 205 0, width=0.9\textwidth]{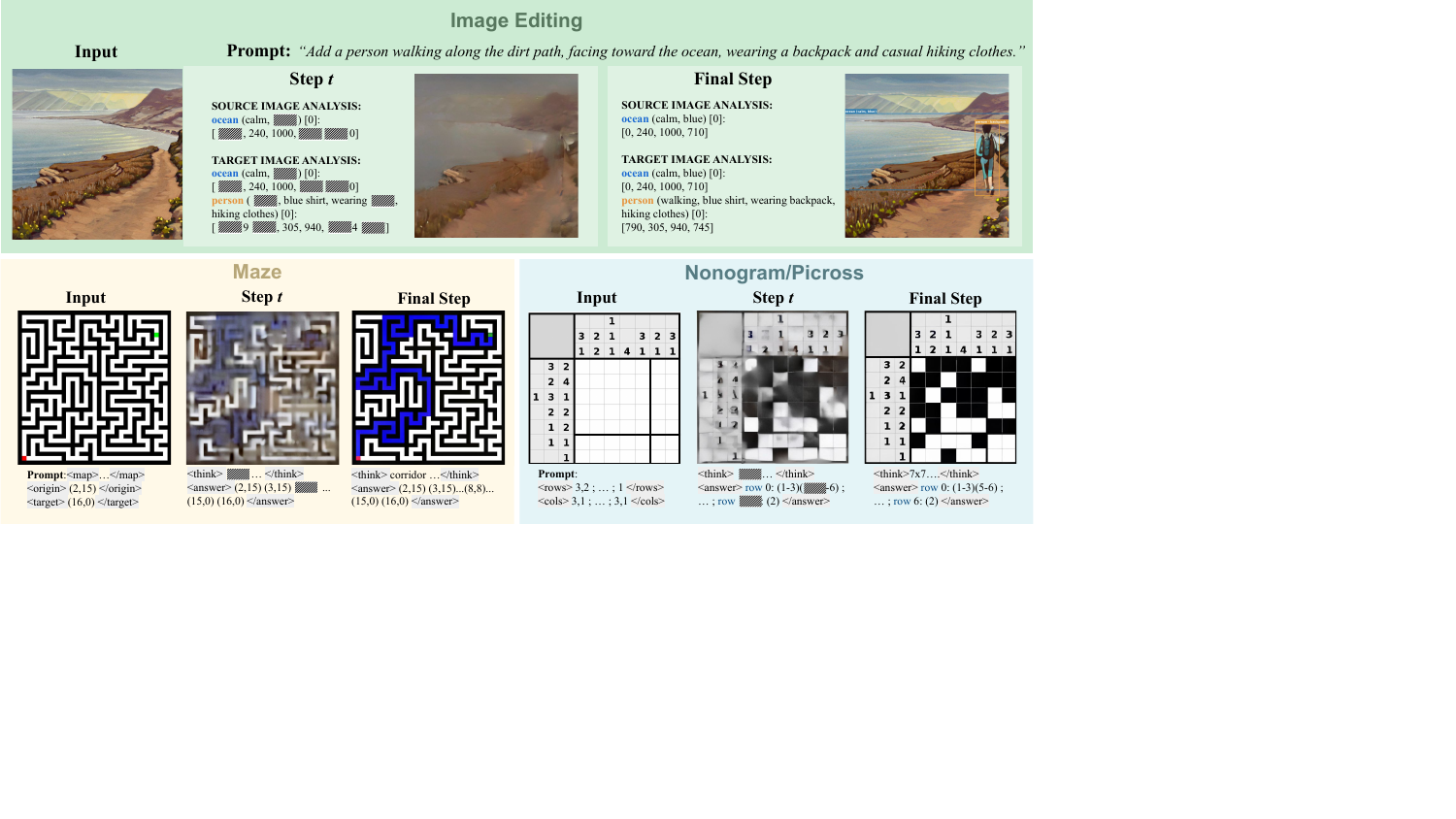}
  \vspace{-2mm}
  \caption{\textbf{\method{} in action: text and image co-author the answer.} Three trajectories of \method{} on image editing, maze, and nonogram solving, showing the joint state at an intermediate step $t$ and at the final step. The image-editing panel highlights the core mechanism: at step $t$ the text branch has already begun committing a \textit{target}-image bounding box in text for the new object \emph{person}; by the final step the image branch has placed the hiker \emph{exactly} inside the finalized box (we overlay the bounding boxes from generated text on edited image). The text branch \emph{plans} where the edit should land, and the image branch \emph{executes} that plan within the same denoising trajectory --- no second forward pass, no external grounder. Maze and Nonogram show the same coupled-refinement pattern: partial-path and partial cell-fill commitments at step $t$ converge with their text-side answers by the final step.}
  \vspace{-20mm}
  \label{fig:teaser}
\end{figure*}



\section{Introduction}
\label{sec:intro}
When a teacher explains an idea at a whiteboard, language and drawing take place \emph{together}: each utterance affects the sketch, and each new mark affects the next sentence. Understanding and generation are not separate stages but a tightly coupled loop, with each modality continuously informing and revising the other as the explanation unfolds. We aim to operationalize this loop for artificial systems, producing text and image content concurrently rather than sequentially, with one modality shaping the other \emph{as} it is generated. While our method is modality-agnostic in principle, in this paper we focus on pairing text for understanding, with images for generation.

Masked Diffusion Models (MDMs)~\citep{austin2021structured, sahoo2024simple, lou2024discrete, shi2024simplified} are well-suited for parallel multimodal generation. Unlike autoregressive sequential pipelines~\citep{deng2025emerging, chen2025janus, xie2026showo}, which first decode the entire textual reasoning trace and then condition image synthesis on it -- a unidirectional flow that cannot retract early reasoning errors -- MDMs predict all masked tokens simultaneously, admit a clean continuous-time formulation as Markov Jump Processes~\citep{campbell2022continuous, berghaus2024foundation}, and scale naturally to multiple modalities under a unified vocabulary spanning text and image tokens~\citep{xin2025lumina, yang2025mmada}. Their parallel structure makes them, in principle, a natural framework for joint  text/image generation in a single decoding loop.

In practice, existing MDM samplers~\citep{sahoo2024simple, wang2026remasking, tian2026mmadaparallel, chen2026unified, ouyang2026training} fall short of \emph{true} concurrent joint multimodal generation. Even samplers that nominally decode both modalities in parallel~\citep{chen2026unified, tian2026mmadaparallel} factorize each denoising step so that the text and image updates depend only on the previous joint state and not \emph{within} the same step; concurrency reduces to interleaving over a shared history. The resulting trajectories might drift: text might commit to descriptions the image cannot illustrate, the image might render content the text never described. Compounding this, standard masked diffusion is unable to remask~\citep{austin2021structured, campbell2022continuous, sahoo2024simple}, once a token is committed it cannot be revised.  Cross-modal contradictions introduced by an uncoupled parallel decoder persist for the rest of sampling.

We address both problems jointly. We introduce \textbf{Self-Correcting Coupled Markov Jump Processes (SC-CMJP)}, a general framework for concurrent joint multimodal generation in which the two modalities actively cross-weigh their commitments \emph{within} every denoising step. One modality's transition rates become functionals of the other modality's unmasking confidence score, weighted by cross-modal attention extracted from the same backbone forward pass.  The unmasking schedule of one modality adapts to the confidence of the other modality. Combined with a remasking jump in the spirit of ReMDM~\citep{wang2026remasking}, this lifts decoding from one-way unmasking to a bidirectional birth-death (unmask-remask) process that can both reveal new tokens and retract earlier ones whenever cross-modal evidence turns against them.

Along with SC-CMJP, we design a single-pass training-free sampler \method{} (Self-\underline{CO}rrecting  \underline{CO}upled  Jump) that highlights two core ideas: \emph{Coupling} the per-modality transition rates through cross-modal attention, and \emph{Correcting} earlier commitments via a remasking jump. \method{} runs on a frozen MDM with no architectural change, no auxiliary evaluator, and a single forward pass per step. Figure~\ref{fig:teaser} illustrates \method{} on all three of our benchmarks; in the image-editing trajectory, the text branch commits a bounding box for the inserted object and the image branch fills it in. In the \emph{same} step, the text \emph{plans} where the edit should land, the image branch \emph{executes} that plan.

To validate \method{}, we instantiate it on three concurrent text-and-image tasks of increasing semantic difficulty: image editing on an extended ImgEditBench~\citep{ye2025imgedit} protocol with mAP-style~\citep{lin2014microsoft} grounded-understanding metrics, and two new visual-reasoning tasks,  a maze and a nonogram (\textsc{JMaze} and \textsc{JNono}) where text and image are logically interlocked and jointly verifiable against algorithmic ground truth. To enable training and evaluation, we curate three corpora: \textsc{JEdit-1M}, \textsc{JMaze-200K}, and \textsc{JNono-200K}, all of which we plan to release. Across all three tasks, \method{} consistently improves both modalities, beats existing sampling methods \citep{sahoo2024simple, wang2026remasking,tian2026mmadaparallel} on concurrent joint image undestanding and generation, espectially joint performance metrics. In addition, \method{} sampler's performance scales monotonically with the number of denoising steps. In summary, our main contributions are:
\begin{itemize}
    \item We introduce the first approach to model simultaneous image understanding and generation as a single, unified stochastic process. To achieve this, we propose Self-Correcting Coupled Markov Jump Processes, that integrate parallel joint multimodal generation with built-in self-correction.
    \item We design a novel coupled sampler \method{} for joint multimodal sampling, running in a single forward pass per step. 
    \item We curated three large-scale joint-generation corpora (\textsc{JEdit-1M}, \textsc{JMaze-200K}, \textsc{JNono-200K}) along with matching benchmarks that probe both in-distribution and out-of-distribution performance. We have a plan to release model checkpoints, code, datasets to community.
    \item \method{} improves the  state-of-the-art  on concurrent image understanding and editing, as well as  visual reasoning tasks against existing sampling methods. Our sampler's performance scales monotonically with the number of denoising steps -- empirical evidence that the benefits of cross-modal coupling \emph{compound} across the trajectory.
\end{itemize}

\section{Background}
\label{sec:background}

\subsection{Masked Diffusion Models and Remasking}
\label{sec:bg_mdm}

Masked discrete diffusion models~\citep{austin2021structured, sahoo2024simple, lou2024discrete, shi2024simplified} corrupt a clean sample $\x \in \mathcal{V}^L$ by gradually replacing tokens with a special absorbing state $\mask$, and learn to invert this corruption. We adopt the continuous-time formulation of \citet{campbell2022continuous} as our default view because it admits the cross-modal coupling and remasking extensions developed in this paper.

\paragraph{Forward process and CTMC equivalence.}
For $t \in [0, 1]$ and a monotonically decreasing noise schedule $\at \in [0, 1]$ with $\alpha_0 \approx 1$, $\alpha_1 \approx 0$, the per-position marginal of the forward process is
\begin{equation}
\label{eq:mdlm_marginal}
q(\z_t \mid \x) = \cat\bigl(\z_t;\ \at\,\x + (1 - \at)\,\mask\bigr),
\end{equation}
factorized across positions. The same dynamics admit an equivalent Continuous-Time Markov Chain (CTMC) description: states evolve by stochastic jumps between $\x$ and $\mask$ at the infinitesimal rate
{\small
\begin{equation}
\label{eq:fwd_rate}
\Rt \;=\; -\frac{\dot\alpha_t}{\at}\bigl(\mathbf{I} - \mask \mathbf{1}^{\!\top}\bigr),
\end{equation}
}
under which the integrated transition probability $q(\z_t = \x \mid \x) = \at$ recovers \Eqn{eq:mdlm_marginal} exactly. \citet{sahoo2024simple} establish that the discrete-time absorbing-state formulation of MDLM and the CTMC formulation of \citet{campbell2022continuous} parameterize the same family of marginals, posteriors, and likelihood bounds; we use the two views interchangeably.

\paragraph{Training objective.}
A denoising network $\x_\theta$ is trained to predict the clean state $\x$ from $\z_t$, and the resulting NELBO collapses to a position-wise weighted cross-entropy~\citep{sahoo2024simple, shi2024simplified}:
\begin{equation}
\label{eq:mdlm_nelbo}
\lossnelbo \;=\; \mathbb{E}_{q,\,t}\!\int_{0}^{1} \frac{\dot\alpha_t}{1 - \at}\sum_{\ell=1}^{L} \log\bigl\langle \x_\theta^{\ell}(\z_t, t),\ \x^{\ell}\bigr\rangle\,\mathrm{d} t.
\end{equation}
The corresponding reverse posterior of standard MDLM has a well-known inability to remask~\citep{austin2021structured, campbell2022continuous, sahoo2024simple}: once a token is unmasked, no subsequent reverse step can revisit it, so any error committed at a timestep $t$ persists for the rest of the trajectory.

\paragraph{Remasking via $\sigma_t$.}
ReMDM~\citep{wang2026remasking} repairs this by allowing committed tokens to revert to $\mask$ with per-step probability $\sigma_t \in [0, 1]$, yielding the modified reverse posterior
{\small
\begin{equation}
\label{eq:remdm_posterior}
q_\sigma(\z_s \mid \z_t, \x) =
\begin{cases}
\cat\!\bigl(\z_s;\ (1 - \sigma_t)\,\x + \sigma_t\,\mask\bigr), & \z_t \neq \mask \\[2pt]
\cat\!\left(\z_s;\ \dfrac{\alpha_s - (1-\sigma_t)\alpha_t}{1 - \alpha_t}\,\x + \dfrac{1 - \alpha_s - \sigma_t \alpha_t}{1 - \alpha_t}\,\mask\right), & \z_t = \mask,
\end{cases}
\end{equation}
}
which preserves the marginal in \Eqn{eq:mdlm_marginal} when $\sigma_t \le \min(1, (1-\alpha_s)/\alpha_t)$ and recovers MDLM at $\sigma_t = 0$. The remask jump makes the forward process non-Markovian, but the reverse stays Markovian and tractable~\citep{wang2026remasking}. Existing $\sigma_t$ schedules~\citep{wang2026remasking} are \emph{modality-agnostic} -- they score remasks from intra-modal likelihoods alone, leaving cross-modal contradictions undetected in joint generation.

\subsection{Markov Jump Processes and Coupled Multimodal Generation}
\label{sec:bg_cmjp}

A Markov Jump Process (MJP) \citep{campbell2022continuous, berghaus2024foundation} is a continuous-time Markov process on a discrete state space specified by a rate matrix $R_t(z, z')$; the CTMC view \citep{campbell2022continuous, sahoo2024simple, ouyang2026training} of MDM is exactly such an MJP over $\mathcal V^L$ with rate matrix \Eqn{eq:fwd_rate}. Exact reverse simulation via Gillespie's algorithm~\citep{gillespie1976general, gillespie1977exact} updates one position per jump and is prohibitive at modern sequence lengths, so practical samplers use $\tau$-leaping~\citep{gillespie2001approximate} -- a parallel approximation that updates all positions simultaneously, of which the standard MDM reverse step is the first-order discretization~\citep{campbell2022continuous}.

\paragraph{Coupled MJPs for Multimodal Generation.}
Coupled jump processes are classical in chemistry~\citep{gillespie1977exact} and Glauber dynamics~\citep{glauber1963time}, but their coupling structure is hand-specified and they target physical simulation; standard MJP-based diffusion likewise treats per-position rates as conditionally independent given $\x_\theta(\z_t)$ even in multimodal settings. We instead define a \emph{Coupled Markov Jump Process} (CMJP) over $\z_t = (\z_t^{\texttt{text}}, \z_t^{\texttt{image}})$ with modality-specific rate matrices $\Rt^{a}$ whose intensities depend on the hidden representations and instantaneous confidence of the complementary modality through learned cross-modal attention. Combined with a ReMDM-style $\sigma_t$ jump (\Eqn{eq:remdm_posterior}), birth (unmask) and death (remask) rates of one modality are informed by the current commitments of the other, enabling localized cross-modal self-correction at sampling time.

\section{Related Work}
\label{sec:related}

\paragraph{Discrete Diffusion.}
Absorbing-state discrete diffusion was introduced in D3PM~\citep{austin2021structured} and refined into score-entropy~\citep{lou2024discrete}, simplified-ELBO~\citep{sahoo2024simple, shi2024simplified}, and any-order autoregressive~\citep{ou2024your} formulations. Scaling to LLMs has been driven by LLaDA~\citep{nie2025large}, Dream~\citep{ye2025dream7d}, and SDAR~\citep{cheng2025sdar}. Decoding improvements include block-wise generation~\citep{arriola2025block}, Top-$K$ confidence selection~\citep{nie2025large, kim2025train}, and length-adaptive scheduling~\citep{ou2024your}. Multimodal extensions tokenize images with VQ-VAE~\citep{oord2017neural} and operate over a unified vocabulary spanning text and image tokens, \emph{e.g.} Lumina-DiMOO~\citep{xin2025lumina} and MMaDA~\citep{yang2025mmada}.

\paragraph{Self-Correction via Remasking.}
Three families of method address the failure-to-remask problem. \emph{Predictor-corrector samplers}~\citep{campbell2022continuous, lezama2023discrete, gat2024discrete, campbell2024generative} reduce $\tau$-leaping error via corrector steps without an explicit remask jump. \emph{Training-based remasking} modifies the model: GIDD~\citep{rutte2025generalized} generalizes forward and reverse processes; \citet{zhao2024informed} train a separate hollow-transformer evaluator; \citet{kim2025fine, huang2025dont} fine-tune the pretrained MDM to estimate per-token quality. \emph{Training-free remasking} keeps the backbone frozen: ReMDM~\citep{wang2026remasking} adds a heuristic $\sigma_t$ schedule, \citet{peng2025path} explore path-following corrections, and \citet{ouyang2026training} use cumulative-confidence signals. \method{} sits within the training-free family, but is the first to couple the remasking signal across modalities and treat cross-modal contradiction as the self-correction trigger.

\paragraph{Concurrent Multimodal Samplers.}
UD-VLA~\citep{chen2026unified} factorizes the joint into independent per-modality terms (an instance of MDM~\citep{sahoo2024simple} sampling); MMaDA-Parallel~\citep{tian2026mmadaparallel} interleaves text and image updates across steps but samples them \emph{independently within each step} -- both modality updates condition only on the previous joint state, with no cross-modal feedback inside the step itself, so any coupling is realized only through shared history rather than instantaneous negotiation. Our experiments compare \method{} against three representative samplers covering these regimes: MDM~\citep{sahoo2024simple}, ReMDM~\citep{wang2026remasking}, and MMaDA-Parallel~\citep{tian2026mmadaparallel}. Coupled jump processes in chemistry and statistical physics~\citep{gillespie1977exact, glauber1963time} use hand-specified couplings for physical simulation, so neither methods transfer to our setting.

\section{Self-Correcting Coupled Markov Jump Processes}
\label{sec:method}
\begin{figure*}[t]
  \centering
  \includegraphics[width=\textwidth]{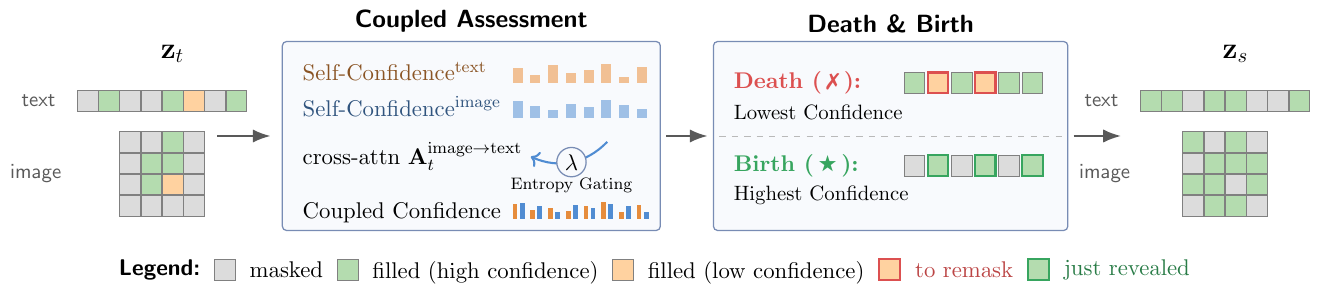}
  \caption{\textbf{\method{} sampler.} A single denoising step from $\z_t$ to $\z_s$. From one forward pass, the model produces per-token Self-Confidence for both modalities; cross-modal attention $\mathbf{A}^{\text{image}\to\text{text}}_t$ propagates text confidence to image positions, and an entropy-based gate $\lambda$ mixes self and cross signals into Coupled Confidence. The \emph{Death} jump remasks the lowest-confidence committed tokens, and the \emph{Birth} jump reveals the highest-confidence masked tokens under the noise schedule.}
  \label{fig:method}
\end{figure*}
Standard multimodal diffusion samplers often suffer from \textit{modality-drift}, where the generated text and image diverge because their unmasking schedules are independent and irreversible. We address this by reformulating joint generation as a unified system where modalities actively negotiate their commitment through a shared birth-death jump process whose intensities are coupled across modalities. Figure~\ref{fig:method} illustrates a single denoising step of the resulting sampler.

\subsection{Continuous-Time Likelihood Bounds for Joint Generation}
\label{sec:method_likelihood}
Following MDM derivation~\citep{sahoo2024simple}, for $t \in [0, 1]$ and a monotonically decreasing noise schedule $\at \in [0, 1]$, the per-position marginal of the forward process is
\begin{equation}
\label{eq:mdm_marginal}
q(\z_t \mid \x) = \cat\bigl(\z_t;\ \at\,\x + (1 - \at)\,\mask\bigr),
\end{equation}
factorized across positions. 
We treat text and image as a single sequence over the union vocabulary $\mathcal{V} = \mathcal{V}_{\texttt{text}} \cup \mathcal{V}_{\texttt{image}}$ and share a single absorbing token $\mask$ across both modalities, so that the per-position marginal in \Eqn{eq:mdm_marginal} applies uniformly to text and image positions; this is what allows the joint NELBO in \Eqn{eq:joint_loss} below to factor cleanly without a cross-modality term. A joint clean sample is $\x = (\x^{\texttt{text}}, \x^{\texttt{image}}) \in \mathcal{V}^L$ with total vocabulary size $L = L^{\texttt{text}} + L^{\texttt{image}}$, and is corrupted by $q(\z_t \mid \x)$, applied position-wise across both modalities under a shared schedule $\at$.

Because the per-position marginal factorizes and both modalities use the same $\at$, the joint NELBO inherits the similar form in MDM~\citep{sahoo2024simple} with no cross-modality term in the objective:
\begin{equation}
\label{eq:joint_loss}
\lossnelbo \;=\; \mathbb{E}_{q,\,t}\!\int_{0}^{1} \frac{\dot\alpha_t}{1-\at}\sum_{\ell=1}^{L} \log\bigl\langle \x_\theta^{\ell}(\z_t, t),\ \x^{\ell}\bigr\rangle\,\mathrm{d}t.
\end{equation}
A crucial property is implicit in the conditioning: the network output $\x_\theta^{\ell}(\z_t, t)$ at any position $\ell$ has access to the \emph{entire} joint state $\z_t = (\z_t^{\texttt{text}}, \z_t^{\texttt{image}})$, including the complementary modality. Training under \Eqn{eq:joint_loss} therefore implicitly forces $\x_\theta$ to learn cross-modal denoising signals -- to reconstruct a masked image patch from surrounding pixels \emph{and} concurrent textual clues, and conversely to reconstruct a masked text token from surrounding context \emph{and} the partially decoded image. The NELBO never references this dependency explicitly, yet at convergence it is encoded in the network's hidden representations and attention patterns. Our coupled sampler \method extracts these latent cross-modal signals at inference time via a single forward pass through $\x_\theta$, with no architectural changes to the trained model and no auxiliary cross-modal evaluator.

\subsection{Asymmetric Sequential Sampling via Chain-Rule Decomposition}
\label{sec:chain_rule}
Let $c$ denote the task conditioning context (e.g., the editing prompt together with the source image for image editing, or the puzzle specification for visual reasoning). A naive sampler for the joint reverse posterior $p_\theta(\z_s^{\texttt{text}}, \z_s^{\texttt{image}} \mid \z_t, c)$ updates both modalities independently from the same $\z_t$, ignoring that the latest text decision $\z_s^{\texttt{text}}$ provides immediate context for inferring the corresponding image state $\z_s^{\texttt{image}}$. We instead exploit the chain-rule factorization
\begin{equation}
\label{eq:chain_rule}
p(\z_s^{\texttt{text}}, \z_s^{\texttt{image}} \mid \z_t, c)
\;=\;
\underbrace{p(\z_s^{\texttt{text}} \mid \z_t, c)}_{\text{(I) text update}}
\cdot
\underbrace{p(\z_s^{\texttt{image}} \mid \z_t, \z_s^{\texttt{text}}, c)}_{\text{(II) image update conditioned on text at step $s$}},
\end{equation}
which makes explicit that the text update needs no information beyond the current $\z_t$, while the image update would ideally condition on the latest sampled $\z_s^{\texttt{text}}$.

Evaluating Term~(II) exactly demands a second forward pass through $\x_\theta$ at every diffusion step -- prohibitive at inference time. We avoid this cost by approximating $\z_s^{\texttt{text}}$ with quantities already produced by the single forward pass at $\z_t$: the image side reads the model's self-belief over $\z_t^{\texttt{text}}$, propagates it through cross-modal attention, and uses the result as a low-cost surrogate for the unavailable $\z_s^{\texttt{text}}$. This yields an \emph{asymmetric scoring rule}: text positions are scored by their pure self-confidence, while image positions are scored by an entropy-gated mixture of self-confidence and a cross-modal signal -- the \textit{Coupled Confidence}. Both scores then drive the death-and-birth dynamics.

\subsection{Self-Confidence and Coupled Confidence}
\label{sec:confidence}

\textbf{Self-Belief via Gumbel-Max.}
At each timestep, the network produces beliefs $p_\theta(\x^\ell \mid \z_t, c)$ over each position's clean token. To escape local minima, we apply Gumbel-max sampling, $\hat{\x}^\ell = \arg\max_{v \in \mathcal{V}} \bigl(\log p_\theta(v \mid \z_t, c) + \gamma_v\bigr)$ with i.i.d.~Gumbel noise $\gamma_v$. The \emph{Self-Confidence} of position $\ell$ is the model's probability for its sampled choice: $\texttt{SelfConf}_\ell = p_\theta(\hat{\x}^\ell \mid \z_t, c)$.
\paragraph{Text Scoring.}
Because Term~(I) of \Eqn{eq:chain_rule} has no cross-modal dependency, the per-position text score is the self-confidence directly: 
\begin{equation}
    \texttt{Score}_{\texttt{text}, \ell} \;=\; \texttt{SelfConf}_{\texttt{text}, \ell}.
\label{eq:text_score}
\end{equation}
\paragraph{Cross-Modal Negotiation.}
For Term~(II), we approximate the conditioning on $\z_s^{\texttt{text}}$ using $\z_t^{\texttt{text}}$ via cross-attention. Let $\mathbf{H}_t^{\texttt{image}} \in \mathbb{R}^{L^{\texttt{image}} \times D}$ and $\mathbf{H}_t^{\texttt{text}} \in \mathbb{R}^{L^{\texttt{text}} \times D}$ be the hidden representations of the two modalities, extracted post-hoc from the model's final layer during the same forward pass that produced $\texttt{SelfConf}$ -- no architectural modification or auxiliary attention head is introduced. We compute the image$\to$text cross-attention
\begin{equation}
\label{eq:cross_attn}
\mathbf{A}_t^{\texttt{image} \to \texttt{text}}
\;=\; \texttt{Softmax}\!\left( \frac{\mathbf{H}_t^{\texttt{image}} (\mathbf{H}_t^{\texttt{text}})^\top}{\sqrt{D}} + \mathbf{B}_t \right),
\end{equation}
where $\mathbf{B}_t$ is a mask-aware bias that downweights attention toward currently masked positions. The cross-modal negotiation is implemented as the cross signal for image position $\ell$, which is the attention-weighted text self-confidence:
\begin{equation}
\label{eq:cross_signal}
\texttt{CrossSignal}_\ell \;=\; \sum_{j=1}^{L^{\texttt{text}}} \mathbf{A}_{t, \ell j}^{\texttt{image} \to \texttt{text}} \cdot \texttt{SelfConf}_{\texttt{text}, j}.
\end{equation}
\paragraph{Dynamic Trust through Entropy-Based Gating.}
The weight given to the cross signal versus the self signal on the image side is a \emph{single scalar gate} $\lambda \in [0, 1]$, computed once per denoising step and applied uniformly to every image position. We derive $\lambda$ from per-modality predictive uncertainty: letting $\bar{\mathcal{H}}_a = \tfrac{1}{L^a}\sum_{\ell=1}^{L^a}\bigl[ -\sum_{v} p_\theta(v \mid \z_t, c)_{a,\ell} \log p_\theta(v \mid \z_t, c)_{a,\ell} \bigr]$ denote the mean token-level Shannon entropy in modality $a$, averaged over \emph{all} positions of that modality,
\begin{equation}
\label{eq:lambda_gate}
\lambda \;=\; \frac{\bar{\mathcal{H}}_{\texttt{image}}}{\bar{\mathcal{H}}_{\texttt{image}} + \bar{\mathcal{H}}_{\texttt{text}} + \epsilon}.
\end{equation}
When the image is locally chaotic (high $\bar{\mathcal{H}}_{\texttt{image}}$), $\lambda \to 1$ and the image modality defers to the text for evidentiary support; when the text is the uncertain side, $\lambda \to 0$ and the image relies on its own self-belief. The single-gate formulation follows directly from \Eqn{eq:chain_rule} -- only the image term carries a cross-modal dependency to be resolved.

\paragraph{Shared Percentile Rank Space.}
Because the text and image vocabularies differ in size ($|\mathcal{V}_{\texttt{text}}| \gg |\mathcal{V}_{\texttt{image}}|$), absolute confidence values are not directly comparable across modalities. We therefore project each scalar score into its empirical percentile rank within its own modality, ensuring the image-side mixture below is well-conditioned regardless of vocabulary scale.

\paragraph{Coupled Confidence for Image Scoring.} Combining the rank-normalized self and cross signals:
{\small
\begin{equation}
\label{eq:coupled_conf}
\texttt{Score}_{\texttt{image}, \ell}
\;=\;
\texttt{CoupledConf}_\ell
\;=\;
(1 - \lambda) \cdot \text{Rank}(\texttt{SelfConf}_{\texttt{image}, \ell})
+ \lambda \cdot \text{Rank}(\texttt{CrossSignal}_\ell).
\end{equation}
}
\subsection{The \method Sampler}
We interpret the joint reverse generation as a modality-specific \emph{birth-death} jump process. Standard masked diffusion only ``births'' tokens (unmask) and never updates them. \method{} additionally allows tokens of either modality to ``die'' (remask) when their scores fall below the schedule-driven death rate, allowing both intra-modal errors and cross-modal contradictions to be retracted.

\paragraph{Death Step (Remasking).}
At each discretization step $(t \to s)$ we set the death rate $\sigma_t$ to satisfy the valid-posterior constraint $\sigma_t \le \min(\eta, (1 - \alpha_s)/\alpha_t)$. For each modality $a$, the idealized unmasked count is $U_t^a = \alpha_t L^a$ and the idealized masked count is $M_t^a = (1-\alpha_t) L^a$ (we use $\lfloor\cdot\rfloor$ 
to obtain integer token counts). The death quota is $N_{\text{remask}}^a = \lfloor U_t^a \cdot \sigma_t \rfloor$, and the $N_{\text{remask}}^a$ unmasked tokens with the \emph{lowest} $\texttt{Score}_a$ are reverted to absorbing token $\mask$. Per the asymmetric rule (\Eqn{eq:chain_rule}), remasking uses $\texttt{SelfConf}$ for text (\Eqn{eq:text_score}) and $\texttt{CoupledConf}$ for image (\Eqn{eq:coupled_conf}).

\paragraph{Birth Step (Unmasking).}
To keep the global signal-to-noise progression matched to the schedule, the birth quota
$N_{\text{unmask}}^a = \lfloor \tfrac{\alpha_s - \alpha_t}{1 - \alpha_t} \cdot M_t^a \rfloor + \Delta N^a$
combines the base unmask count with an extra-unmask compensation $\Delta N^a = \lfloor \sigma_t \cdot \tfrac{\alpha_t}{1-\alpha_t} \cdot M_t^a \rfloor$ derived from the ReMDM \citep{wang2026remasking} posterior. Among the post-remask masked tokens of modality $a$, we reveal the $N_{\text{unmask}}^a$ with the \emph{highest} $\texttt{Score}_a$, again with the asymmetric definition.  Algorithm~\ref{alg:sccmjp} gives the full pseudocode.

\begin{algorithm}[H]
\caption{\method{} Sampler (Asymmetric Scoring)}
\label{alg:sccmjp}
\begin{algorithmic}[1]
\STATE {\bfseries Input:} Pretrained $\x_\theta$, schedule $\alpha_t$, remasking ratio $\eta$, steps $T$, modality lengths $L^{\texttt{text}}, L^{\texttt{image}}$, conditioning $c$.
\STATE Initialize $\z_1 \leftarrow \{\mask\}^{L^{\texttt{text}} + L^{\texttt{image}}}$.
\FOR{$i = T$ {\bfseries down to} $1$}
    \STATE $t \leftarrow i/T,\ \ s \leftarrow (i-1)/T$
    \STATE $\z_s \leftarrow \z_t$ \hfill // initialize next-step state; death/birth jumps below write into $\z_s$
    \STATE \textbf{// Single forward pass}
    \STATE Run $\x_\theta(\z_t, c)$; extract $\mathbf{H}_t^{\texttt{text}}, \mathbf{H}_t^{\texttt{image}}$ and mean entropies $\bar{\mathcal{H}}_{\texttt{text}}, \bar{\mathcal{H}}_{\texttt{image}}$ over all positions
    \STATE Sample $\hat{\x}^\ell$ via Gumbel noise; record $\texttt{SelfConf}_\ell$ for both modalities
    \STATE \textbf{// Asymmetric scoring (\Eqn{eq:chain_rule})}
    \STATE \emph{Text} \ \ \,$\texttt{Score}_{\texttt{text}, \ell} \leftarrow \texttt{SelfConf}_{\texttt{text}, \ell}$ \hfill // Term~(I): no cross-modal dependence
    \STATE \emph{Image} compute $\mathbf{A}_t^{\texttt{image}\to\texttt{text}}$, $\texttt{CrossSignal}_\ell$ (Eqs.~\ref{eq:cross_attn} and \ref{eq:cross_signal}), scalar gate $\lambda$ (Eq.~\ref{eq:lambda_gate})
    \STATE \quad\ \ \ $\texttt{Score}_{\texttt{image}, \ell} \leftarrow \texttt{CoupledConf}_\ell$ via \Eqn{eq:coupled_conf} \hfill // Term~(II): approx.\ $\z_s^{\texttt{text}}$ via attention
    \STATE \textbf{// Death \& Birth Jumps}
    \STATE $\sigma_t \leftarrow \min\!\bigl(\eta,\ (1-\alpha_s)/\alpha_t\bigr)$ \hfill // valid-posterior constraint
    \FOR{$a \in \{\texttt{text}, \texttt{image}\}$}
        \STATE $U_t^a \leftarrow \lfloor \alpha_t L^a \rfloor$,\ \ $M_t^a \leftarrow \lfloor (1-\alpha_t) L^a \rfloor$ \hfill // idealized counts, floored at runtime
        \STATE $N_{\text{remask}}^a \leftarrow \lfloor U_t^a \sigma_t \rfloor$,\ \ $N_{\text{unmask}}^a \leftarrow \bigl\lfloor \tfrac{\alpha_s - \alpha_t}{1-\alpha_t}\, M_t^a \bigr\rfloor + \bigl\lfloor \sigma_t\, \tfrac{\alpha_t}{1-\alpha_t}\, M_t^a \bigr\rfloor$
        \STATE \emph{Death:} in $\z_s$, remask the $N_{\text{remask}}^a$ tokens of modality $a$ with \emph{lowest} $\texttt{Score}_a$ to $\mask$
        \STATE \emph{Birth:} in $\z_s$, reveal the $N_{\text{unmask}}^a$ masked tokens of modality $a$ with \emph{highest} $\texttt{Score}_a$
    \ENDFOR
\ENDFOR
\STATE {\bfseries Output:} Unified multimodal sample $\z_0$
\end{algorithmic}
\end{algorithm}

\section{Datasets and Benchmarks for Joint Multimodal Generation}
\label{sec:datasets}
\begin{figure}[t]
    \centering
    \includegraphics[trim=0 420 1020 0, clip, width=\textwidth]{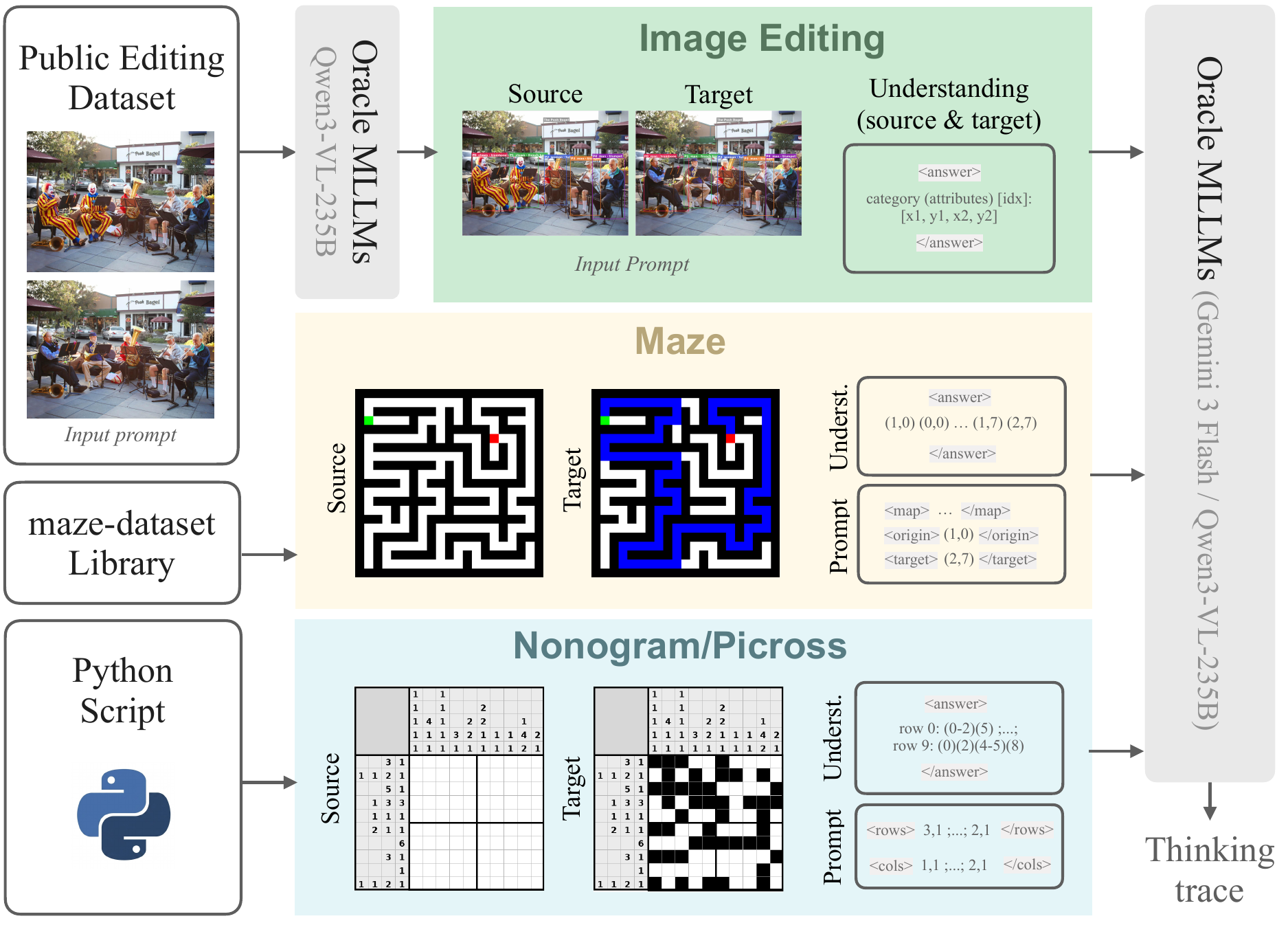}
    \caption{\textbf{Dataset curation pipeline.} For \textsc{JEdit-1M}, raw editing pairs are augmented by an oracle Qwen3-VL-235B that produces both the per-image scene-graph understanding and the thinking trace. For \textsc{JMaze-200K} and \textsc{JNono-200K}, source/target images and the structured understanding are produced algorithmically.}
    \label{fig:dataset_curation}
\end{figure}

To train and evaluate \method{} across both photographic and logical
multimodal tasks, we curate three joint-generation datasets: \textsc{JEdit-1M} for image editing, \textsc{JMaze-200K} for maze solving, and \textsc{JNono-200K} for nonogram solving, each paired
with a held-out benchmark. All three corpora share the same record
schema (prompt, source image, target image, structured understanding,
thinking trace); Figure~\ref{fig:dataset_curation} shows the curation pipeline.

\subsection{\textsc{JEdit-1M}: Joint Image Editing and Understanding}
\label{sec:datasets_imgedit}

\begin{figure}[t]
    \centering
    \includegraphics[trim=0 430 1160 0, clip, width=\textwidth]{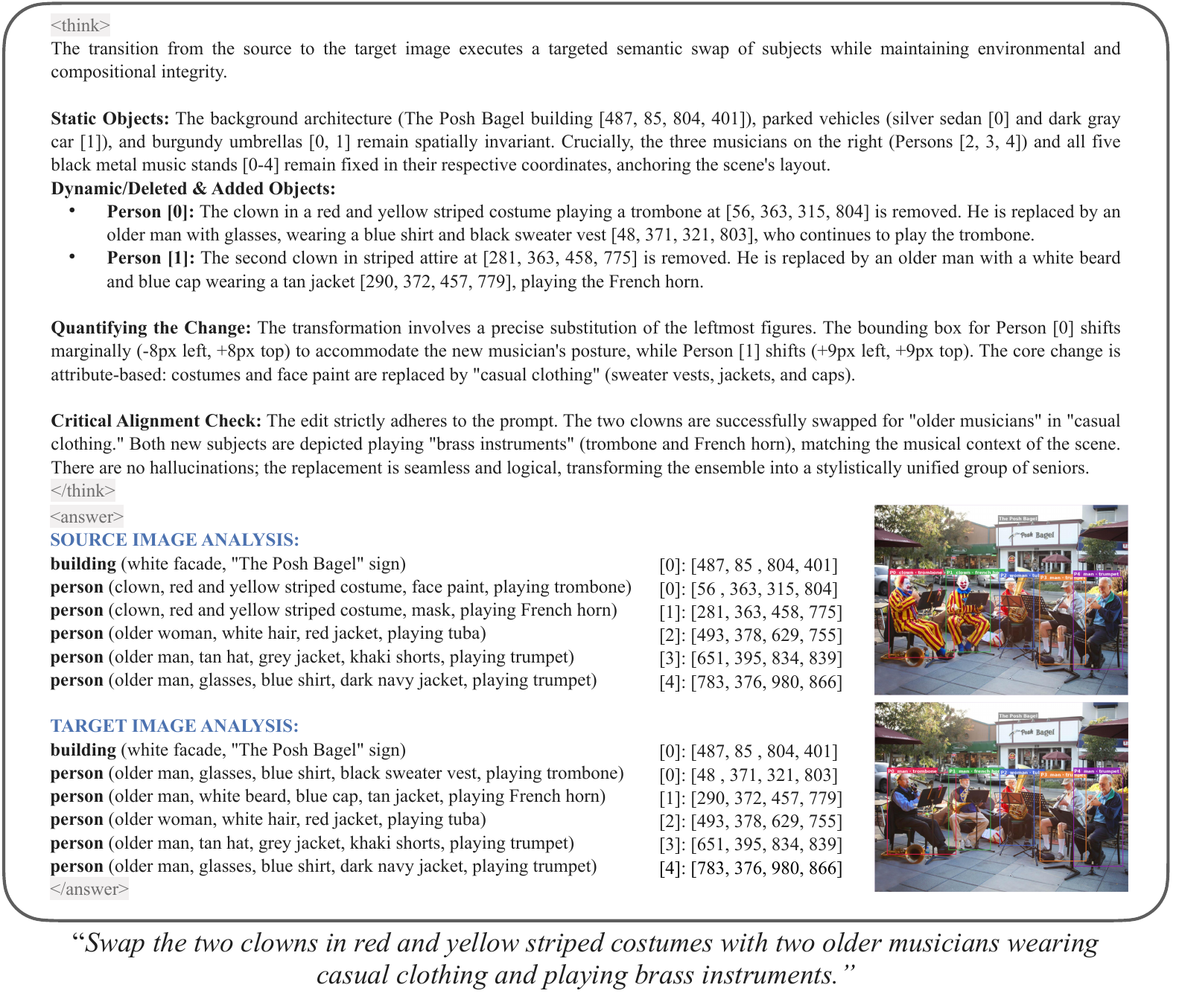}
    \vspace{-2mm}
    \caption{\textbf{\textsc{JEdit-1M} training sample.} A natural-language edit prompt (bottom) paired with the source and target images (right), the oracle Qwen3-VL-235B thinking trace, and the per-panel source/target scene-graph understanding. The bounding boxes and labels from MLLM's grounding solution are overlaid on the corresponding images.}
    \label{fig:dataset_sample}
    \vspace{-2mm}
\end{figure}

\paragraph{Training corpus.}
\textsc{JEdit-1M} combines a 724k-pair subset of
ImgEdit~\citep{ye2025imgedit} with a 368k-pair subset of
OmniEdit~\citep{wei2025omniedit}, yielding 1M (prompt, source, target)
tuples. Because raw editing data lacks the joint
understanding and thinking supervision that joint multimodal generation requires, we augment
each pair with two fields synthesized by
Qwen3-VL-235B~\citep{bai2025qwen3vltechnicalreport}: (i)~a pixel-aligned
source/target scene-graph extraction with per-category indices preserved
across panels, and (ii)~a logic-based thinking trace conditioned on both
images, the prompt, and the scene graph from~(i). Figure~\ref{fig:dataset_sample} provides a concrete example of a training sample of the \textsc{JEdit-1M} dataset.

\paragraph{Benchmark and metrics.}
We extend ImgEditBench~\citep{ye2025imgedit} with two metric families
that score joint understanding\,+\,generation behavior. \textbf{Text
quality} is generative perplexity under a frozen GPT-2
Large~\citep{radford2019language} plus token-distribution entropy as a
mode-collapse safeguard, following MDLM~\citep{sahoo2024simple} and ReMDM~\citep{wang2026remasking}.
\textbf{Image-edit fidelity} is the standard ImgEditBench oracle score
from Gemini~3~Flash~\citep{gemini3flash}. 
\textbf{Image-grounded understanding} is COCO-style mAP@0.5:0.95~\citep{lin2014microsoft} against a \emph{pseudo} scene
graph constructed per-sample by Gemini on the model's own (source,
generated-target) pair: pseudo grounding is necessary because each
model produces a different target image. We report mAP separately for
source and target sections plus their average.

\subsection{\textsc{JMaze-200K} and \textsc{JNono-200K}: Visual Reasoning}
\label{sec:datasets_maze_nonogram}

To stress-test joint generation on tasks where text and image are
\emph{logically interlocked}, we add two synthetic visual-reasoning
corpora. In both, the input fully specifies a unique solution and the
two modalities are independently verifiable against algorithmic ground
truth.

\paragraph{Corpora.}
We use the $\texttt{maze-dataset}$~\citep{ivanitskiy2023mazedataset} library
to generate \textsc{JMaze-200K} comprising 200k DFS-perfect mazes with grid sizes
uniformly sampled from $\{6,\ldots,20\}$; the input shows walls plus a
green start and red end, the output overlays the solution path in blue,
and the text answer is the path as an $(r,c)$ sequence.
\textsc{JNono-200K} is 200k nonogram puzzles with grid sizes
$\{5,\ldots,25\}$, generated by a mixture of pattern synthesizers
biased toward multi-run row/column clues; the input renders the empty
$N\!\times\!N$ grid with marginal clue numbers, and the output fills
cells black to satisfy every clue. For both corpora, the thinking trace
is synthesized by an oracle Gemini~3~Flash conditioned on the source/target images and the algorithmic
ground-truth answer.

\paragraph{Parallel-form supervision.}
Both the structured understanding and the thinking trace are written
in a form native to parallel decoding rather than autoregressive
ordering: maze solutions use global $(r,c)$ coordinates instead of
relative moves (\emph{Left/Right/Up/Down}), and nonogram supervision
emphasizes bidirectional row/column constraint propagation instead of
row-by-row solving. A relative direction is only well-defined once the
previous coordinate is committed, while a global coordinate keeps its
meaning regardless of which other tokens have been unmasked, aligning
the supervision with the parallel denoising the sampler actually
performs.

\paragraph{Benchmarks and metrics.}
For each task we hold out a 500-sample test set whose grid sizes
\emph{exceed} the training range on both ends:
\textsc{JMaze-Test500} spans $\{3,\ldots,22\}$
(in-dist $\{6,\ldots,20\}$, OOD $\{3,4,5,21,22\}$);
\textsc{JNono-Test500} spans $\{3,\ldots,27\}$
(in-dist $\{5,\ldots,25\}$, OOD $\{3,4,26,27\}$). Our headline metric
is \textbf{joint accuracy}, a sample counts correct only when both
the text answer (judged by a tolerant Gemini extractor against the
algorithmic ground truth) and the generated image (compared
side-by-side with the ground-truth image, by Gemini~3~Flash) are
correct. \textbf{Text accuracy} and \textbf{image accuracy} are
reported separately as diagnostics; on in-distribution
and OOD splits.

\section{Experiments}
\label{sec:experiments}
We evaluate \method{} against three sampler families across the three
tasks of Section~\ref{sec:datasets}. To ensure a strictly controlled comparison:
we initialize from Lumina-DiMOO~\citep{xin2025lumina}, fine-tune on the
corresponding corpus, and swap only the inference-time sampler. For
each task, we fine-tune the model on $64 \times$ H100 80\,GB GPUs with a
total batch size of $512$ and learning rate $2\!\times\!10^{-5}$.
Baselines includes: \textbf{MDM}~\citep{sahoo2024simple}, independent modalities with no
remasking; \textbf{ReMDM}~\citep{wang2026remasking}, independent modalities
with remasking; and
\textbf{MMaDA-Parallel}~\citep{tian2026mmadaparallel}, modalities interleave and are independent with no remasking. \method{} is
the only entry that generates joint modalities concurrently with
cross-modal coupling and self-correction. All samplers use a cosine $\alpha_t$ schedule; for the remasking variants
(ReMDM and \method{}), we use a remasking schedule with $\sigma_t = 0.01$
on $t \in [0.25, 0.75]$ and $0$ elsewhere.

\subsection{Joint Image Editing and Understanding}
\label{sec:exp_imgedit}

\begin{table*}[t]
\centering
\caption{\textbf{Image-editing results on the extended ImgEditBench
protocol.} Bottom block: apple-to-apple sampler comparison; above the
rule: pre-trained Lumina-DiMOO and Qwen3-VL-8B references.
\textbf{Bold}/\underline{underline} = best/second best across the four
fine-tuned samplers.}
\label{tab:imgedit}
\resizebox{\textwidth}{!}{%
\begin{tabular}{l c c c c c c c}
\toprule
& & \multicolumn{2}{c}{Text Generation} & Image Generation & \multicolumn{3}{c}{Image Understanding (mAP@0.5:0.95 $\uparrow$)} \\
\cmidrule(lr){3-4} \cmidrule(lr){5-5} \cmidrule(lr){6-8}
Method & \# turns $\downarrow$ &
GPT-2L PPL $\downarrow$ & Entropy $\uparrow$ &
ImgEditBench $\uparrow$ &
Source & Target & Overall \\
\midrule
Qwen3-VL-8B               & 2 (I2I $\to$ I2T)  & 11.9128            & 4.7523          & --            & 0.391          & 0.330          & 0.360          \\
Lumina-DiMOO    & 3 (I2I $+$ 2$\times$I2T) & 3.1647             & 3.0513          & 1.68          & 0.019          & 0.019          & 0.019          \\
\midrule
MDM                                           & 1                        & 8.5780             & \underline{4.5160} & 1.78       & 0.386          & 0.322          & 0.354          \\
ReMDM                                         & 1                        & \textbf{8.4863}    & 4.4929          & 1.73          & 0.388          & 0.318          & 0.353          \\
MMaDA-Parallel                                & 1                        & 8.8363             & \textbf{4.5353} & 1.44          & 0.366          & 0.304          & 0.335          \\
\rowcolor{oursfill}\method{} \textbf{(Ours)}                              & 1                        & \underline{8.5649} & 4.5005          & \textbf{1.93} & \textbf{0.392} & \textbf{0.346} & \textbf{0.369} \\
\bottomrule
\end{tabular}}
\end{table*}

\begin{figure*}[t]
  \centering
  \includegraphics[width=\textwidth]{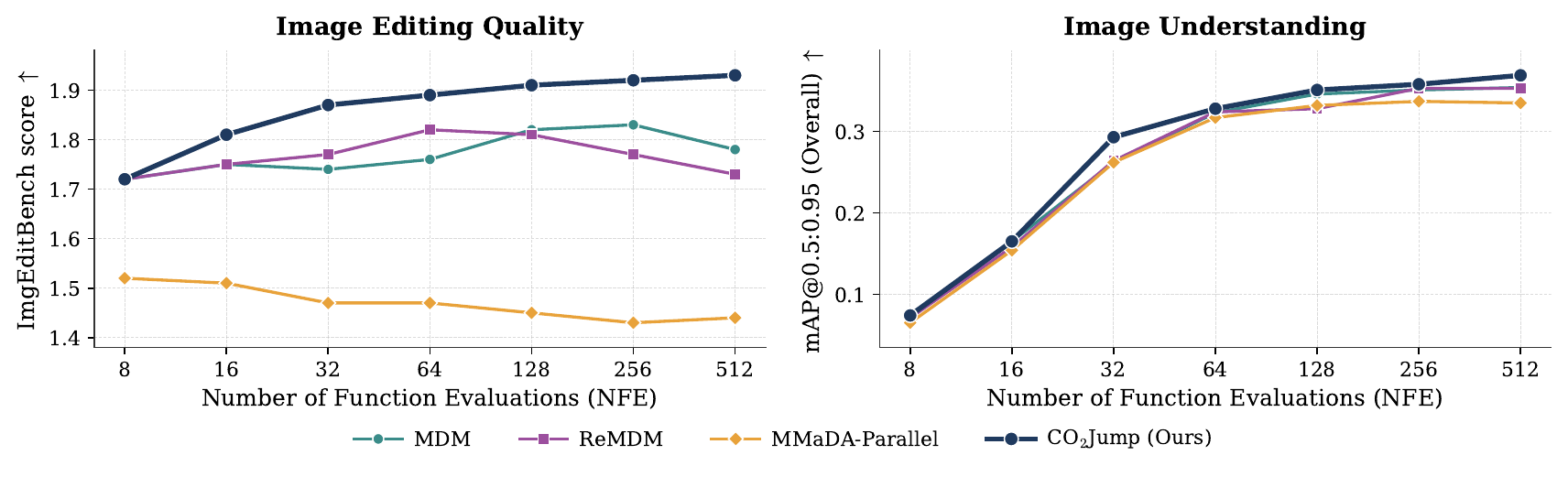}
  \caption{\textbf{Scaling sampling steps.} ImgEditBench (left) and
  overall mAP (right) vs.\ NFE. \method{} is the only sampler with monotonically rising curves; the gap to baselines widens with NFE.}
  \label{fig:scaling_nfe}
\end{figure*}

Table~\ref{tab:imgedit} reports results on extended ImgEditBench. Our
headline metric is per-section mAP@0.5:0.95 from the pseudo-grounding
pipeline; target-mAP is the cleanest test of cross-modal coupling
because it requires the image branch to place each edited object
inside the bounding boxes that the text branch declares \emph{and} the text
branch to describe what the image branch actually produces.

\paragraph{Coupled generation beats a strong sequential grounder.}
Qwen3-VL-8B is a stringent reference: same scale as our backbone, and
the supervision used to fine-tune all four samplers comes from the
Qwen3-VL family \citep{bai2025qwen3vltechnicalreport}. Because Qwen3-VL-8B itself cannot generate images, we
evaluate its grounding on the same (source, generated-target) image pairs
produced by our \method{} sampler, effectively giving it the generated
target image as input for free. Yet \method{} produces grounded text
\emph{concurrently} with the image and still exceeds Qwen3-VL-8B on the
same images, on both target mAP ($0.346$ vs.\ $0.330$) and overall mAP
($0.369$ vs.\ $0.360$).

\paragraph{Better understanding leads to better generation.} Pre-trained
Lumina-DiMOO has near-zero understanding mAP ($0.019$); fine-tuning
unlocks it across all samplers, and the sampler with the strongest
understanding (\method{}, $0.369$ overall) also achieves the best
ImgEditBench score ($1.93$), while the weakest (MMaDA-Parallel,
$0.335$) is worst on ImgEditBench ($1.44$). This is the empirical
signature of the chain-rule decomposition
(Section~\ref{sec:chain_rule}): once text decisions inform image
decisions within a single step, sharper text grounding tightens the
image-side score distribution.


\paragraph{Scaling sampling steps.} Sweeping NFE from 8 to 512
(Figure~\ref{fig:scaling_nfe}), \method{} is the only sampler that
improves \emph{monotonically} on both metrics: ImgEditBench rises from
$1.72$ to $1.93$ and overall mAP from $0.074$ to $0.369$.
Single-modality baselines plateau and regress at high NFE, and
MMaDA-Parallel \emph{degrades} from $1.52$ to $1.44$, its uncoupled
schedule cannot productively use additional steps. The gap also
widens: at 8 NFE the four methods sit within $0.009$ mAP, but at 512
NFE \method{} is $+0.015 / +0.016 / +0.034$ ahead of MDM / ReMDM /
MMaDA-Parallel. Coupling \emph{compounds} across steps rather than
saturating.

\subsection{Visual Reasoning: Maze and Nonogram Solving}
\label{sec:exp_reasoning}

Table~\ref{tab:reasoning} reports joint accuracy on
visual reasoning benchmarks on in-distribution and OOD grid
sizes.

\paragraph{\method{} outperforms other samplers in all six columns.} Improvements over the best
baseline range from $+0.008$ (Maze, In-Dist) to $+0.062$ (Nonogram,
OOD). No baseline is the runner-up everywhere --- MDM is second on
Maze but fourth on Nonogram, and MMaDA-Parallel is the reverse --- so
consistent joint correctness across both task structures is itself
evidence that the coupling mechanism is generic rather than
benchmark-specific.
\begin{table*}[t]
\centering
\caption{\textbf{Joint accuracy on \textsc{JMaze-Test500} and
\textsc{JNono-Test500}.} A sample is correct only when text and image
both match the algorithmic ground truth.
\textbf{Bold}/\underline{underline} = best/second best.}
\label{tab:reasoning}
\setlength{\tabcolsep}{6pt}
\resizebox{\textwidth}{!}{%
\begin{tabular}{l c c c c c c}
\toprule
& \multicolumn{3}{c}{\cellcolor{mazefill}Maze Solving (\textsc{JMaze-Test500})}
& \multicolumn{3}{c}{\cellcolor{nonofill}Nonogram Solving (\textsc{JNono-Test500})} \\
\cmidrule(lr){2-4} \cmidrule(lr){5-7}
Method
 & \shortstack{In-Distribution\\$6{\times}6\!-\!20{\times}20$}
 & \shortstack{OOD\\$\{3,4,5,21,22\}^2$}
 & \shortstack{Total\\$3{\times}3\!-\!22{\times}22$}
 & \shortstack{In-Distribution\\$5{\times}5\!-\!25{\times}25$}
 & \shortstack{OOD\\$\{3,4,26,27\}^2$}
 & \shortstack{Total\\$3{\times}3\!-\!27{\times}27$} \\
\midrule
MDM              & \underline{0.461} & \underline{0.312} & \underline{0.424} & 0.110             & 0.050             & 0.100             \\
ReMDM            & 0.229             & 0.032             & 0.180             & 0.002             & 0.000             & 0.002             \\
MMaDA-Parallel   & 0.419             & 0.304             & 0.390             & \underline{0.143} & \underline{0.113} & \underline{0.138} \\
\cellcolor{oursfill}\textbf{\method{} (Ours)}
 & \cellcolor{oursfill}\textbf{0.469}
 & \cellcolor{oursfill}\textbf{0.320}
 & \cellcolor{oursfill}\textbf{0.432}
 & \cellcolor{oursfill}\textbf{0.167}
 & \cellcolor{oursfill}\textbf{0.175}
 & \cellcolor{oursfill}\textbf{0.168} \\
\bottomrule
\end{tabular}}
\end{table*}

\begin{figure*}[t]
  \centering
  \includegraphics[width=\textwidth]{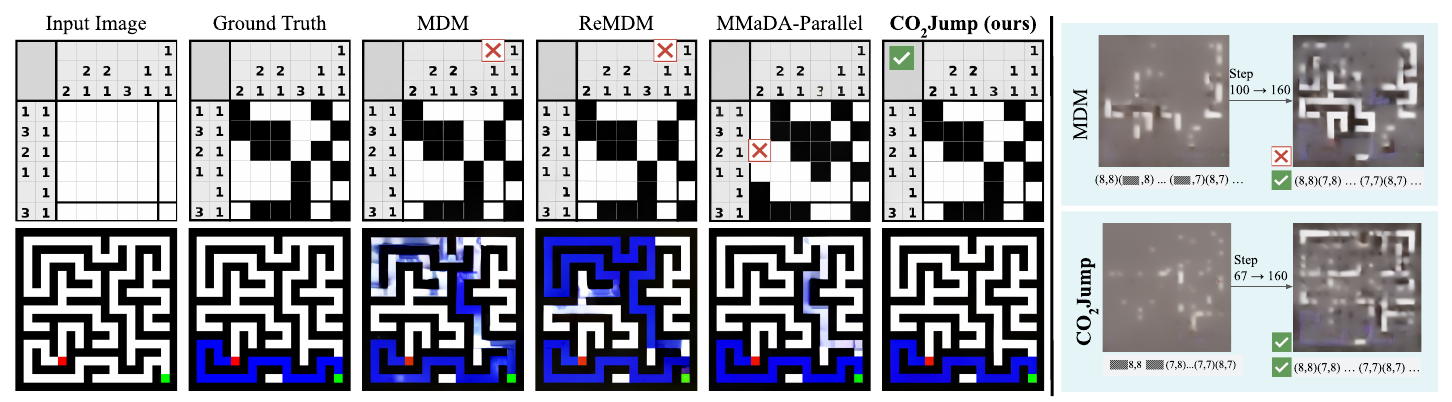}
  \caption{\textbf{Qualitative comparison.} Top left: Nonogram (red
  \textcolor{red}{\ding{55}} marks clue violations); Bottom left: Maze;
  Right: per-step MDM vs.\ \method{} trajectory. Only \method{}
  satisfies all clues and paths, and its joint trajectory stays
  consistent across both modalities.}
  \label{fig:qualitative_reasoning}
\end{figure*}
\paragraph{Out-of-distribution generalization.} \method{} also holds up best on grid sizes outside the training range. On Maze it leads on the OOD split ($0.320$ vs.\ $0.312$ for the runner-up), and on Nonogram it widens its lead substantially ($0.175$ vs.\ $0.113$). MDM drops $55\%$ from In-Dist to OOD on Nonogram and MMaDA-Parallel $21\%$, while \method{} stays roughly flat. It shows that cross-modal coupling transfers solution-style information across grid sizes more robustly than uncoupled samplers.

\paragraph{Qualitative trajectories expose the uncoupling failure mode.}
In Figure~\ref{fig:qualitative_reasoning}, every baseline violates at
least one Nonogram clue (red \textcolor{red}{\ding{55}}), while
\method{} satisfies all clues; on the maze, baselines wander into
wrong corridors or cut through walls and \method{} reproduces the
ground-truth path. The right column exposes \emph{why}: by step 160 of
MDM's trajectory the text answer is correct but the image has drifted
onto a different path, because under MDM's independence factorization
the image branch never sees the latest committed text. \method{}'s
chain-rule decomposition propagates each text commitment into image
scoring at the same step, so both modalities converge on the same
solution.

\subsection{Ablation Study and Analysis}
\label{sec:exp_ablation}

\paragraph{Ablation.} Table~\ref{tab:ablation} removes each mechanism in
isolation. Removing Shared Percentile Rank causes the largest drop in
image-edit fidelity ($1.93 \to 1.87$): without a common scale, one of
Self-Confidence and Cross Signal arbitrarily dominates the blend.
Removing Entropy-Based Gating leaves source-mAP unchanged but cuts
target-mAP by $0.030$ and overall mAP by $0.015$, the gate's main
role is in late-stage sampling where the target image is committed
under freshly grounded text. Removing Self-Correction is the most
damaging change for understanding ($-3.2$ on overall mAP) and drops
editing by $0.01$: without remasking, late-discovered
cross-modal contradictions cannot be repaired.

\begin{table*}[t]
\centering
\caption{\textbf{Ablation on \method{}.} Each row removes one
mechanism: Shared Percentile Rank, Entropy-Based Gating
($\lambda_{\text{image}}$), or Self-Correction.
\textbf{Bold}/\underline{underline} = best/second best.}
\label{tab:ablation}
\setlength{\tabcolsep}{6pt}
\resizebox{\textwidth}{!}{%
\begin{tabular}{l c c c c c c}
\toprule
& \multicolumn{2}{c}{Text Generation} & Image Generation
& \multicolumn{3}{c}{Image Understanding (mAP@0.5:0.95 $\uparrow$)} \\
\cmidrule(lr){2-3} \cmidrule(lr){4-4} \cmidrule(lr){5-7}
Method
 & GPT-2L PPL $\downarrow$ & Entropy $\uparrow$
 & ImgEditBench $\uparrow$
 & Source & Target & Overall \\
\midrule
\cellcolor{oursfill}\textbf{\method{} (full)}
 & \cellcolor{oursfill}\textbf{8.5649}
 & \cellcolor{oursfill}\underline{4.5005}
 & \cellcolor{oursfill}\textbf{1.93}
 & \cellcolor{oursfill}\textbf{0.392}
 & \cellcolor{oursfill}\textbf{0.346}
 & \cellcolor{oursfill}\textbf{0.369} \\
$-$\,Shared Percentile Rank & 8.7117             & 4.4957          & 1.87             & \underline{0.374}          & 0.315             & 0.344             \\
$-$\,Entropy-Based Gating   & \underline{8.7078} & 4.4984          & 1.91             & \textbf{0.392} & \underline{0.316} & \underline{0.354} \\
$-$\,Self-Correction        & 8.8299             & \textbf{4.5180} & \underline{1.92} & 0.363          & 0.310             & 0.337             \\
\bottomrule
\end{tabular}}
\end{table*}

\begin{figure*}[t]
  \centering
  \includegraphics[width=\textwidth]{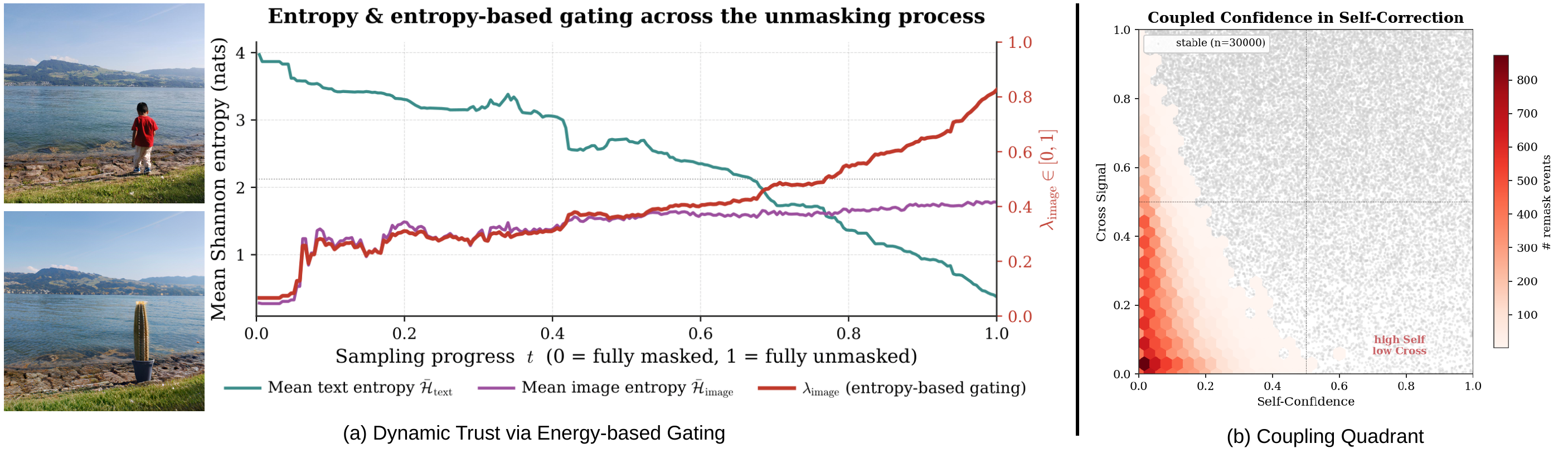}
  \caption{\textbf{(a)} Per-step text/image entropy and the gate
  $\lambda_{\text{image}}$. The image starts confident (source prior),
  text starts uncertain; $\lambda_{\text{image}}$ rises from $\approx 0.05$
  to $\approx 0.85$ as text commits. \textbf{(b)} Image-side remask events
  in (Self-Confidence, Cross Signal) space: most fire in the lower-left
  region; a smaller hotspot at high-Self / low-Cross corresponds to
  coupling-driven revocations.}
  \label{fig:gating_quadrant}
  \vspace{-2mm}
\end{figure*}

\paragraph{Dynamic trust via entropy-based gating.}
Figure~\ref{fig:gating_quadrant} (a) traces per-step entropies and
$\lambda_{\text{image}}$ for a representative editing sample. The
image branch starts confident (source-image prior) while text starts
uncertain (only the editing prompt as context), so
$\lambda_{\text{image}} \approx 0.05$ and the image relies more on
self-confidence. As text commits, its entropy falls and the image
moves into regions outside its prior; $\lambda_{\text{image}}$ rises
to $\approx 0.85$, and the image branch increasingly leans on the
text-grounded cross signal --- the empirical signature of the
chain-rule decomposition.

\paragraph{Coupling quadrant.}
Figure~\ref{fig:gating_quadrant} (b) shows every image-side remask
event in (Self-Confidence, Cross Signal) space. Most events
concentrate in the lower-left region where both signals are low, so
the coupled confidence $(1-\lambda)\cdot\text{Self} +
\lambda\cdot\text{Cross}$ is small and the death jump is triggered. A smaller
hotspot in the high-Self / low-Cross quadrant corresponds to
coupling-driven unmasking --- the image was locally confident but
the text disagreed --- and these account for the gap between the full
\method{} and the ``$-$\,Self-Correction'' ablation row in
Table~\ref{tab:ablation}.

\section{Conclusion}
\label{sec:conclusion}
We presented \textbf{Self-Correcting Coupled Markov Jump Processes (SC-CMJP)}, a framework in which the two modalities of a unified MDM negotiate their commitments \emph{within} every denoising step, and \method{}, a training-free single-pass sampler that instantiates it on a frozen backbone. By coupling one modality's transition rates to the other's emerging confidence via cross-modal attention, and by allowing committed tokens to be retracted through a remasking jump triggered by cross-modal contradictions, \method{} closes the concurrent-generation loop that prior parallel samplers leave open. \method{} delivers best performance across joint image understanding and editing, as well as visual reasoning tasks and it scales monotonically with denoising steps, evidence that cross-modal coupling compounds across the trajectory. Our experiments instantiate the framework on the canonical text-and-image pair; the SC-CMJP recipe is modality-agnostic, and extending it to audio, video, or structured outputs is left to future work.

\newpage
\appendix

\section{Qualitative Example from \method: Joint Image Editing and Understanding}
\label{sec:app_qualitative_imgedit}

\begin{figure}[h]
    \centering
    \includegraphics[width=\textwidth]{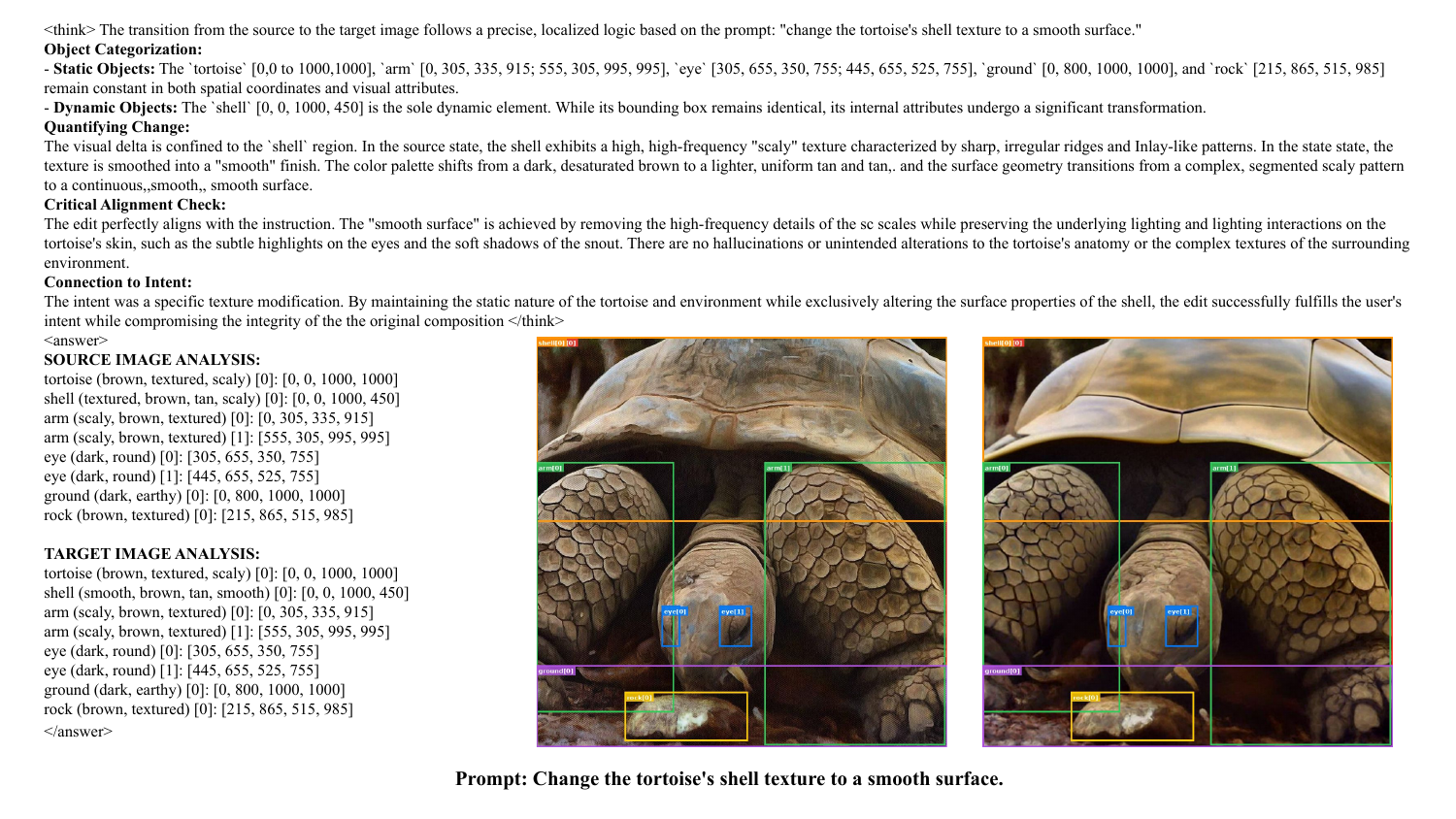}
    \caption{\textbf{Joint image editing with concurrent understanding on \textsc{JEdit-1M}.} Prompt, model-generated thinking trace, scene-graph source/target analyses, and the source/generated images with predicted bounding boxes overlaid (boxes are the model's own text-side predictions, not ground truth). The text branch declares only the \texttt{shell} as a dynamic object --- every other entity has identical bounding boxes in the source and target analyses --- and the image branch realizes exactly this localized texture edit: scaly $\to$ smooth on the shell, with body, limbs, ground and rock preserved.}
    \label{fig:qual_tortoise_imgedit}
\end{figure}

Figure~\ref{fig:qual_tortoise_imgedit} shows a complete \method{} sample from \textsc{ImgEditBench}: prompt, source image, generated target image, the per-panel scene-graph understanding, and the thinking trace, all produced concurrently in a single denoising loop. Bounding boxes overlaid on each image are exactly those emitted by the model's own text branch (no external grounder), normalized to a $1001\!\times\!1001$ canvas. The example illustrates the planning-then-execution pattern that drives the gain on extended ImgEditBench: the text branch isolates \texttt{shell} as the only dynamic object and keeps every other entity (\texttt{tortoise}, \texttt{arm}, \texttt{eye}, \texttt{ground}, \texttt{rock}) at identical bounding boxes between source and target, while changing only the shell's attributes from \texttt{(scaly)} to \texttt{(smooth)}; the image branch realizes precisely that delta, removing the high-frequency scaly pattern on the shell while leaving body, limbs, and surroundings untouched.

\newpage
\section{Qualitative Example from \method: Maze Solving}
\label{sec:app_qualitative_maze}

\begin{figure}[h]
    \centering
    \includegraphics[width=\textwidth]{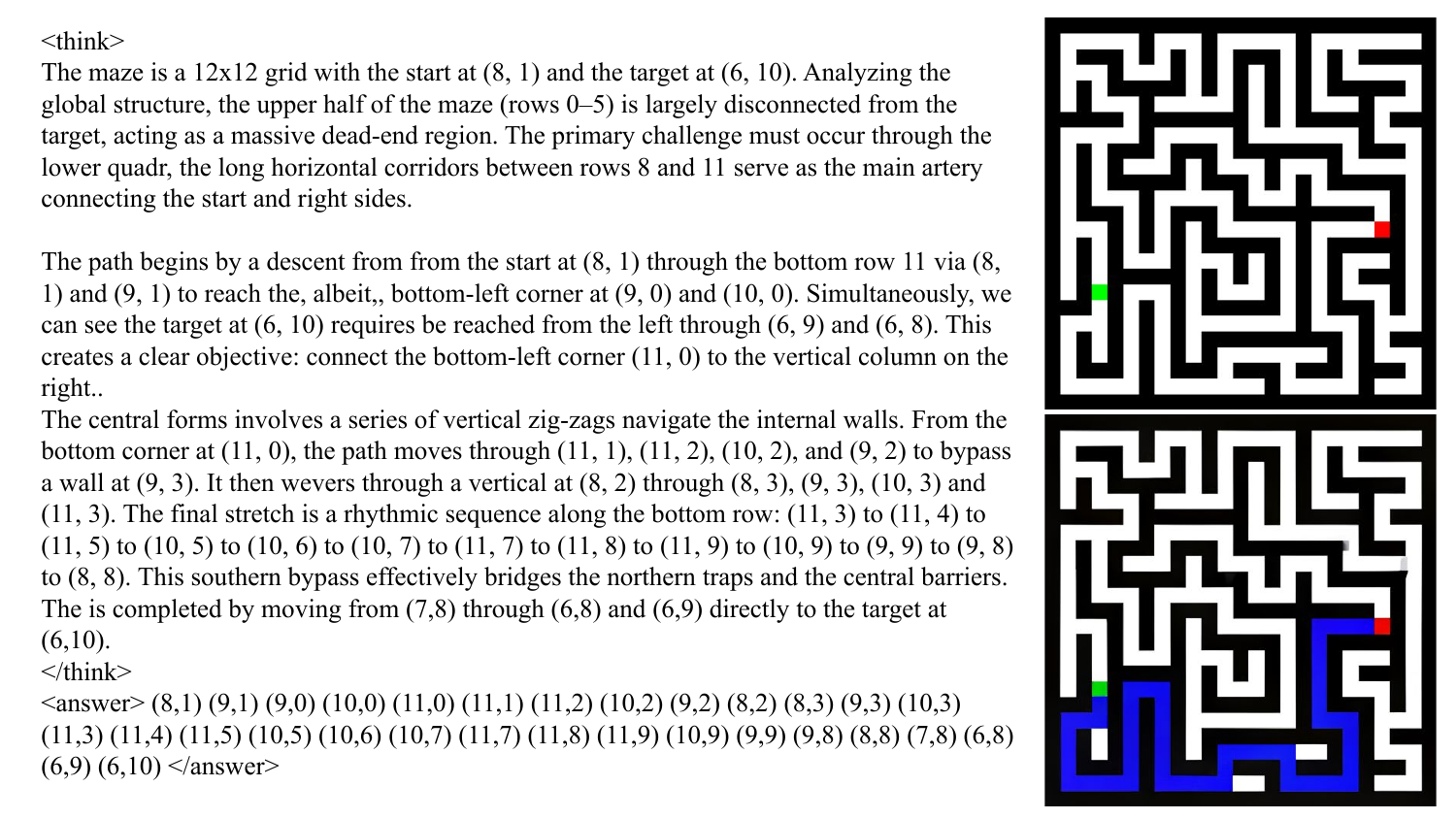}
    \caption{\textbf{Joint maze solving on \textsc{JMaze}.} \method{}'s thinking trace and answer (left) alongside the input maze (top right; green start, red end) and the solved maze with the blue path overlaid (bottom right). The text branch first reasons globally about which corridors are dead-ends and which form the main artery, then commits the path as an explicit $(r,c)$ coordinate sequence; the image branch traces precisely that sequence on the grid in the same denoising trajectory.}
    \label{fig:qual_maze}
\end{figure}

Figure~\ref{fig:qual_maze} shows a \method{} sample from \textsc{JMaze}: the input maze (with green start and red end), the model-generated thinking trace and coordinate path, and the solved maze with the blue path overlaid. Both the textual coordinate sequence and the rendered solution path are produced concurrently in a single denoising loop. The thinking trace reasons globally over the maze --- identifying the upper half as a dead-end region, the lower corridors as the main artery, and the central zig-zag through the bottom row as the southern bypass --- before committing the explicit $(r, c)$ sequence in \texttt{\textless answer\textgreater}, and the image branch traces exactly that sequence, demonstrating that the path drawn on the grid agrees position-by-position with the coordinates the text branch declares.

\newpage
\section{Qualitative Example from \method: Nonogram Solving}
\label{sec:app_qualitative_nonogram}

\begin{figure}[h]
    \centering
    \includegraphics[width=\textwidth]{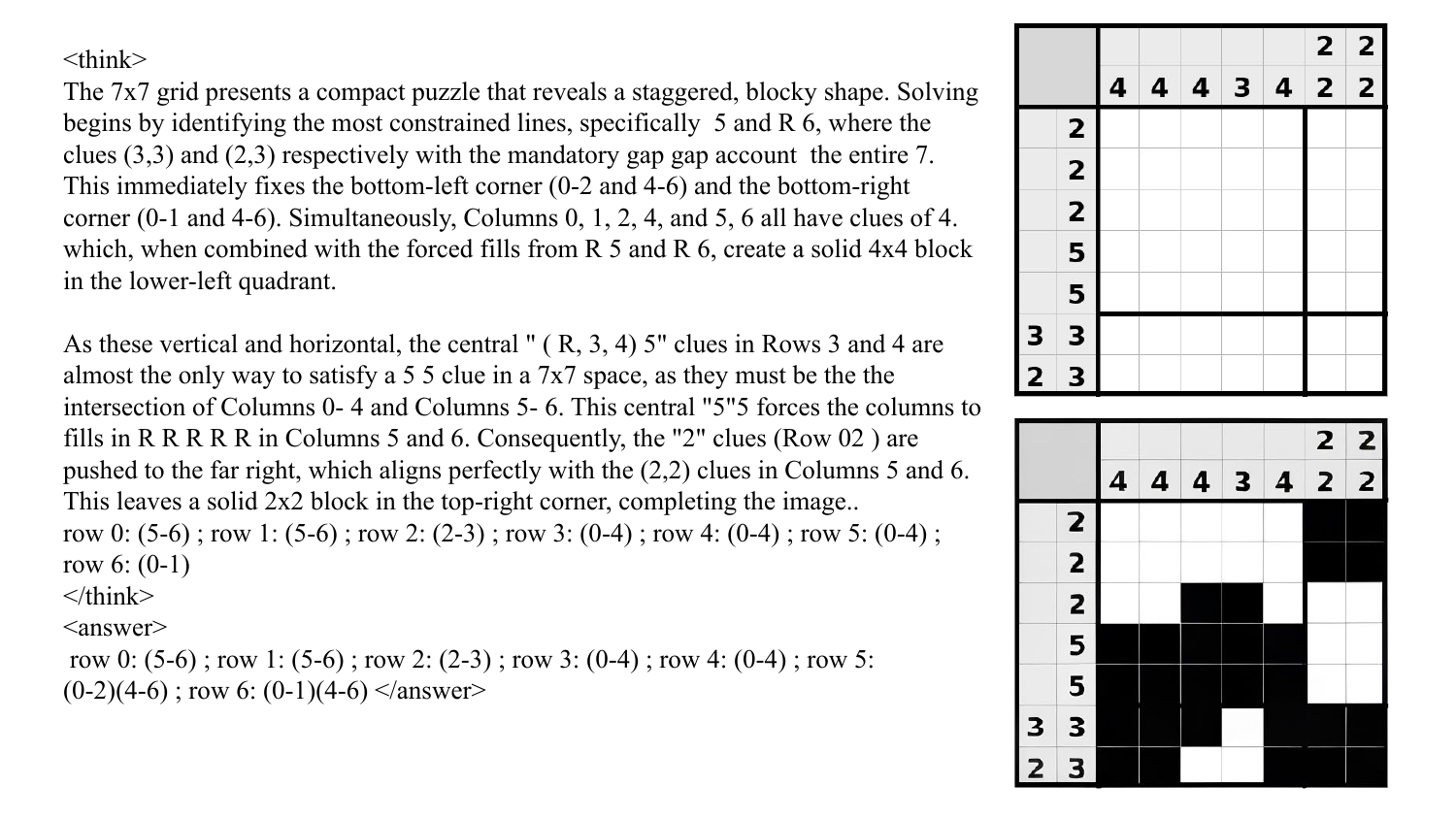}
    \caption{\textbf{Joint nonogram solving on \textsc{JNono}.} \method{}'s thinking trace and answer (left) alongside the input puzzle (top right; row clues at left margin, column clues at top margin) and the solved grid with cells filled black (bottom right). The text branch reasons in a bidirectional, constraint-propagation style --- starting from the most globally-constrained rows and columns and using each forced fill to tighten the other dimension --- before committing the per-row filled-cell ranges; the image branch fills exactly those ranges on the grid in the same denoising trajectory, satisfying every clue.}
    \label{fig:qual_nonogram}
\end{figure}

Figure~\ref{fig:qual_nonogram} shows a \method{} sample from \textsc{JNono}: the input nonogram (an empty $7\!\times\!7$ grid with row and column clue numbers), the model-generated thinking trace and per-row filled-cell answer, and the solved grid with the corresponding cells filled black. The thinking trace performs bidirectional constraint propagation rather than row-by-row solving: it identifies the most globally-constraining lines first (rows whose largest clue forces a 4-cell block, columns whose clue sums match the grid width), explains how the row and column constraints tighten each other concurrently, and only then commits the row-by-row filled-cell ranges in \texttt{\textless answer\textgreater}; the image branch fills precisely those ranges on the grid, satisfying every clue.

\section{System Prompts for Dataset Curation}
\label{sec:app_prompts}

\subsection{Understanding Prompt: Pixel-Aligned Scene Graph Extraction}
\label{sec:app_prompt_underst}

The following system prompt was used to extract the structured understanding (bounding boxes and attributes) from the source and target images.

\begin{center}
\fbox{
\begin{minipage}{0.95\textwidth}
\ttfamily\scriptsize
Task: Pixel-Aligned Scene Graph Extraction for Image Editing Pairs.\\
\\
\#\#\# Objective:\\
Analyze the transition from the Source Image to the Target Image. You\\
must extract a high-fidelity grounding dataset that maps how visual\\
tokens change in response to the Editing Prompt. Focus on maintaining\\
spatial "anchors" for static regions while precisely detailing the\\
"delta" in edited regions.\\
\\
\#\#\# Inputs:\\
1. Source Image: (The original state)\\
2. Target Image: (The modified state)\\
3. Editing Prompt: "\{editing\_prompt\_var\}"\\
\\
\#\#\# Strict Constraints:\\
\textbf{THINKING BEFORE ANSWERING:} In your \textless think\textgreater{} tags, you MUST\\
explicitly compare the two images. List:\\
1. Static entities (Background/Context).\\
2. Transformed entities (Attribute/State changes).\\
3. New or Deleted entities (Structural changes).\\
\\
\textbf{FIDELITY OVER INTENT:} Describe what is visually rendered in the\\
Target Image. If the prompt asks for a "red car" but the Target Image\\
contains a "pink car," you MUST label it as "pink car."\\
\\
\textbf{COMPLETENESS:} Annotate the main subject, all objects involved in\\
the edit, and key background anchors (e.g., floor, sky, walls) to\\
provide global context.\\
\\
\textbf{COORDINATES:} Use normalized integer bounding boxes [x1, y1, x2, y2]\\
on a [0-1000] scale. Bounding boxes must be tight.\\
\\
\textbf{IDENTITY CONSISTENCY:} For any entity present in both images, the\\
'category\_name' and '[index]' MUST be identical.\\
\\
\textbf{PER-CATEGORY INDEXING:} Every new object category MUST start its\\
index at [0]. For example, the first 'car' is 'car [0]', and the\\
first 'person' is 'person [0]'. Do NOT use a single global\\
incrementing index for the whole scene.\\
\\
\textbf{STRUCTURAL LOGIC:} If an object is deleted in the Target, do not\\
list it in the TARGET IMAGE ANALYSIS. If an object is added, list it\\
only in the TARGET section.\\
\\
\#\#\# Prohibited Actions:\\
\textbf{DO NOT} use generic placeholder strings (e.g., "category\_name",\\
"attribute\_list").\\
\textbf{DO NOT} include the brackets around the category name or attributes\\
in the final answer; only the index should remain in brackets as per\\
format.\\
\\
\#\#\# Output Format:\\
\textless answer\textgreater\\
SOURCE IMAGE ANALYSIS:\\
category\_name (attribute\_list) [index]: [x1, y1, x2, y2]\\
\\
TARGET IMAGE ANALYSIS:\\
category\_name (attribute\_list) [index]: [x1, y1, x2, y2]\\
\textless /answer\textgreater\\
\\
\#\#\# Reference Example:\\
\textless answer\textgreater\\
SOURCE IMAGE ANALYSIS:\\
mountain (snowy, jagged) [0]: [0, 50, 1000, 450]\\
person (black hoodie, walking) [0]: [420, 550, 580, 910]\\
person (blue jeans, standing) [1]: [600, 550, 700, 910]\\
\\
TARGET IMAGE ANALYSIS:\\
mountain (snowy, jagged) [0]: [0, 50, 1000, 450]\\
person (yellow raincoat, walking) [0]: [420, 550, 580, 910]\\
person (blue jeans, standing) [1]: [600, 550, 700, 910]\\
\textless /answer\textgreater
\end{minipage}
}
\end{center}

\subsection{\textsc{JEdit-1M} Thinking Prompt: Reasoning Trace Synthesis}
\label{sec:app_prompt_think}

The following system prompt was used with Qwen3-VL-235B to synthesize the logic-based reasoning trace for the \textsc{JEdit-1M} dataset.

\begin{center}
\fbox{
\begin{minipage}{0.95\textwidth}
\ttfamily\scriptsize
Task: Synthesize a Logic-Based Reasoning Trace for Image Editing.\\
\\
\#\#\# Inputs:\\
1. Source and Target Images: Visual evidence.\\
2. Editing Prompt (The intended transformation): \{editing\_prompt\_var\}\\
3. Image Understanding (Bounding box coordinates and object labels\\
for both states): \{answer\}\\
\\
\#\#\# Objective:\\
Synthesize a logic-based reasoning trace that analyzes the visual\\
transition relative to the Editing Prompt. Critically evaluate\\
whether the target state successfully fulfills the prompt's\\
instructions or deviates from the intended logic.\\
\\
\#\#\# Strict Reasoning Guidelines:\\
\textbf{Categorize Objects:} Group objects into: Static (persistent\\
coordinates/attributes), Dynamic (modified coordinates/attributes),\\
and Deleted/Added (present in only one state).\\
\\
\textbf{Quantify Change:} Reference bounding box shifts or attribute\\
updates as direct evidence of the transition.\\
\\
\textbf{Critical Alignment Check:} Explicitly compare the visual delta\\
against the Editing Prompt. Identify hallucinations or instruction\\
mismatches.\\
\\
\textbf{Connect to Intent:} Explain how the detections fulfill—or fail\\
to fulfill—the specific requirements of the Editing Prompt.\\
\\
\textbf{Constraint:} Keep the response dense and information-rich, within\\
a max of 350 words. Avoid generic phrases. Start with the direct\\
logical flow.\\
\\
\#\#\# Output Format:\\
\textless think\textgreater{} [Your concise logical trace here] \textless /think\textgreater
\end{minipage}
}
\end{center}

\subsection{Thinking Prompt: Maze Solving Parallel Reasoning}
\label{sec:app_prompt_maze_think}

The following system prompt was used to synthesize the parallel-reasoning trace for the maze-solving problem in the \textsc{JMaze-200K} dataset.

\begin{center}
\fbox{
\begin{minipage}{0.95\textwidth}
\ttfamily\scriptsize
Task: Synthesize a parallel-reasoning trace for a maze-solving problem.\\
\\
\#\#\# Inputs:\\
1. Input maze image: walls, a green start cell, and a red end cell. No path drawn.\\
2. Output maze image: the same maze with the correct solution path drawn in blue.\\
3. Maze structure as text (adjacency list + start + end):\\
\{prompt\}\\
4. Ground-truth solution path:\\
\{answer\}\\
\\
\#\#\# Objective:\\
Write a natural reasoning trace that reflects a global-to-local refinement process, illustrating how the maze is solved simultaneously rather than purely sequentially.\\
Briefly describe the maze boundaries (grid size, start position, end position).\\
Identify the global structure first: note major bottlenecks, large dead-end regions to mask out, or key anchor points/corridors between the start and end.\\
Describe the path formation as a concurrent process—e.g., establishing a\\
bridge through a central corridor while simultaneously connecting the start\\
and end regions to that main artery, verifying local wall-openings via the\\
adjacency list.\\
Write in natural prose, as a holistic chain of thought. No bullet lists.\\
Keep it under 350 words.\\
\\
\#\#\# Output format (strict — respond with exactly this tag and nothing else):\\
\textless think\textgreater\\
\{\}[Your reasoning trace]\\
\textless /think\textgreater
\end{minipage}
}
\end{center}

\subsection{Thinking Prompt: Nonogram Parallel Reasoning}
\label{sec:app_prompt_nono_think}

The following system prompt was used to synthesize the parallel-reasoning trace for the nonogram puzzle in the \textsc{JNono-200K} dataset.

\begin{center}
\fbox{
\begin{minipage}{0.95\textwidth}
\ttfamily\scriptsize
Task: Synthesize a parallel-reasoning trace for a nonogram (picross) puzzle.\\
\\
\#\#\# Inputs:\\
1. Source image: empty NxN grid with row clue numbers along the left margin and column clue numbers along the top margin.\\
2. Target image: the same grid with cells filled to satisfy every clue, revealing a pixel-art picture.\\
3. Puzzle structure as text (size + row clues + column clues):\\
\{prompt\}\\
4. Ground-truth solution (filled cell ranges per row):\\
\{answer\}\\
\\
\#\#\# Objective:\\
Write a holistic reasoning trace that reflects parallel constraint propagation,\\
NOT row-by-row sequential solving.\\
Briefly describe the puzzle (grid size, rough shape of the revealed picture).\\
Identify the most globally-constraining lines first: rows or columns where\\
  the sum of clues plus required gaps equals N (full-line forces), or lines\\
  whose single largest clue produces "definite-fill" cells via the\\
  left-most / right-most overlap argument.\\
Describe how column constraints and row constraints tighten each other\\
  concurrently: a forced fill produced by a row clue confirms a position in\\
  some column's clue list, which can in turn unlock more rows.\\
Avoid sequential "first solve row 0, then row 1" prose. Emphasize the\\
  bidirectional, global interplay between row and column constraints.\\
Write in natural prose, no bullet lists. Keep it under 250 words.\\
\\
\#\#\# Output format (strict — respond with exactly this tag and nothing else):\\
\textless think\textgreater\\
\{\}[Your reasoning trace]\\
\textless /think\textgreater
\end{minipage}
}
\end{center}

\bibliography{main}

\end{document}